\documentclass[journal]{IEEEtranTIE}
\usepackage{graphicx}
\usepackage{cite}
\usepackage{picinpar}
\usepackage{amsmath}
\usepackage{url}
\usepackage{flushend}   
\usepackage[latin1]{inputenc}
\usepackage{colortbl}
\usepackage{soul}
\usepackage{multirow}
\usepackage{pifont}
\usepackage{color}
\usepackage{alltt}
\usepackage[hidelinks]{hyperref}
\usepackage{enumerate}
\usepackage{siunitx}
\usepackage{breakurl}
\usepackage{epstopdf}
\usepackage{pbox}

\newcommand{\revision}[1]{\textcolor{black}{#1}} 
\newcommand{\secondrevision}[1]{\textcolor{black}{#1}} 
\newcommand{\thirdrevision}[1]{\textcolor{black}{#1}}

\begin{document}
\title{Autonomous Robotic Screening of Tubular Structures based only on\\ Real-Time Ultrasound Imaging Feedback}

\author{Zhongliang Jiang$^{1}$, Zhenyu Li$^{1}$, Matthias Grimm$^{1}$, Mingchuan Zhou$^{2}$, \\Marco Esposito$^{3}$, Wolfgang Wein$^{3}$, Walter Stechele$^{4}$, Thomas Wendler$^{1}$, and Nassir Navab$^{1,5}$
\thanks{Manuscript received Feb 02, 2021; revised Apr 20, 2021, May 17,
2021 and Jun 20, 2021; accepted Jun 22, 2021. \textit{(Zhongliang Jiang and Zhenyu Li contributed equally to this work). (Corresponding author:
Mingchuan Zhou.)}}
\thanks{$^{1}$Z. Jiang, Z. Li, M. Grimm, T. Wendler and N. Navab are with the Chair for Computer Aided Medical Procedures and Augmented Reality, Technical University of Munich, M\"unchen, Germany. {\tt\footnotesize{(zl.jiang@tum.de)}}
        }%
\thanks{$^{2}$M. Zhou is with the College of Biosystems of
Engineering and Food Science, Zhejiang University, China.}
\thanks{$^{3}$M. Esposito and W. Wein are with ImFusion GmbH, Germany.}
\thanks{$^{4}$W. Stechele is with the Chair of Integrated Systems, Technical University of Munich, M\"unchen, Germany.}
\thanks{$^{5}$N. Navab is also with the Laboratory for Computer Aided Medical Procedures, Johns Hopkins University, Baltimore, USA.}
}

\maketitle
	
\begin{abstract} 
Ultrasound (US) imaging is widely employed for diagnosis and staging of vascular  diseases, mainly due to its high availability and the fact it does not emit \revision{ionizing} radiation. 
However, high inter-operator variability \revision{limits the repeatability of US image acquisition.}
To address this challenge, we propose an end-to-end workflow for automatic robotic US screening of tubular structures using only real-time US imaging feedback.
\revision{First,} a U-Net \revision{was trained} for real-time segmentation of vascular structure from cross-sectional US images. 
Then, we represented the detected vascular structure as a 3D point cloud, \revision{which was used} to estimate the centerline of the target structure and its local radius by solving a constrained non-linear optimization problem.
Iterating the previous processes, the US probe was automatically aligned to the normal \revision{direction of} the target structure \revision{while the object \revision{was} constantly maintained in the center of the US view.}
The real-time segmentation result was evaluated both on a phantom and in-vivo on brachial arteries of volunteers. 
In addition, the whole process was validated \revision{using} both simulation and physical phantoms.
\revision{The mean absolute orientation, centering and radius error ($\pm$ SD) on a gel phantom \revision{were} $3.7\pm1.6^{\circ}$, $0.2\pm0.2~mm$ and $0.8\pm 0.4~mm$, respectively.}
The results indicate that the method \revision{can} automatically screen tubular structures with an optimal probe orientation (i.e., normal to the vessel) and accurately estimate the radius of the target structure.
\end{abstract}


\markboth{IEEE TRANSACTIONS ON INDUSTRIAL ELECTRONICS}%
{}

\definecolor{limegreen}{rgb}{0.2, 0.8, 0.2}
\definecolor{forestgreen}{rgb}{0.13, 0.55, 0.13}
\definecolor{greenhtml}{rgb}{0.0, 0.5, 0.0}

\section{Introduction}
\begin{figure}[ht!]
\centering
\includegraphics[width=0.46\textwidth]{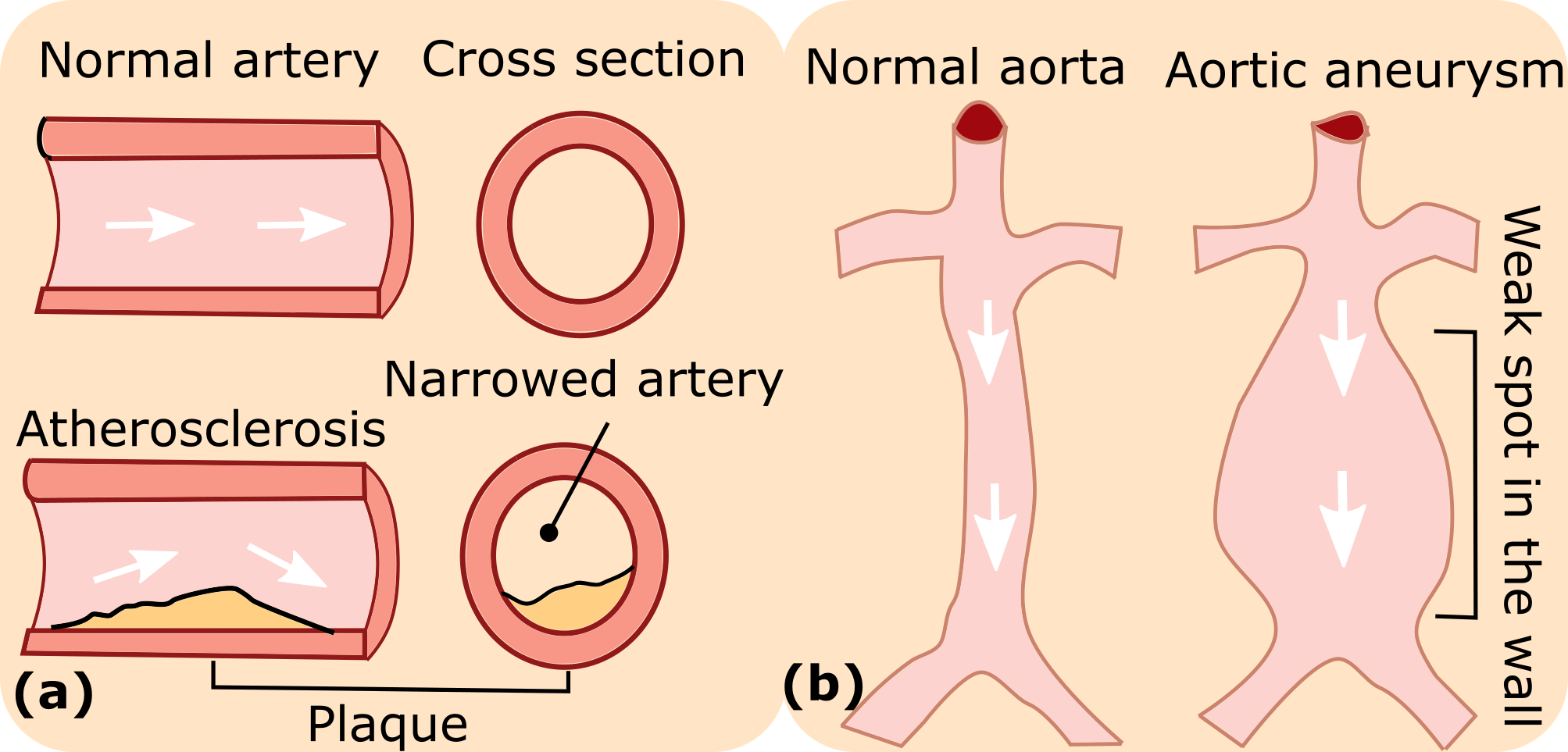}
\caption{\revision{Schematic illustration of the} two most common types of vascular diseases: (a) atherosclerosis and (b) arterial aneurysm.   
}
\label{Fig_background}
\end{figure}

\IEEEPARstart{P}{ERIPHERAL} vascular diseases (PVD) are \revision{some} of the most common diseases, particularly for \revision{older} people.
PVD affects about $20\%$ of adults older than $55$ years and an estimated $27$ million people in North America and Europe~\cite{hankey2006medical}. In the worst case, PVD can lead to critical organ failure, limb amputation, strokes, or heart attacks.
The two most common types of vascular diseases are Atherosclerosis and arterial aneurysm, both \revision{of which are} depicted in Fig.~\ref{Fig_background}. 
\revision{Atherosclerosis is a disease in which plaque builds up inside arteries, whereas an aneurysm refers to a weakening of an artery wall resulting in a bulge or distention of the artery.}
The most common way to diagnose atherosclerosis or arterial aneurysms today is either by measuring the ankle-brachial index (ABI) or by inspecting a computed tomography angiography (CTA) scan. The drawback of \revision{CTA is} that the scans expose both the patient and medical staff to ionizing radiation. \revision{Conversely,} ABI despite being radiation-free, \revision{is unable to provide the location of the PVD} and is highly user-dependent~\cite{stein2006limitation}.


\par
B-mode ultrasound (US) is a promising alternative to CTA for the diagnosis of vascular diseases due to its lack of radiation~\cite{wong2013duplex}. Furthermore, US is widely accessible and cheap, \revision{making} it popular for real-time diagnosis of internal tissues in clinical practice. 
Davis~\emph{et al.} found that conducting US scans for elderly men can reduce premature death from ruptured abdominal aortic aneurysm\revision{s} by up to $50\%$~\cite{davis2013implementation}. Therefore, regular US examination is important for patients. However, the imaging quality \revision{of US} is highly dependent on acquisition parameters (i.e., \revision{the} contact force and probe orientation), which are hard to accurately control in traditional free-hand US, even for experienced sonographers~\cite{yoon2011interobserver}. 
\revision{This leads to a lack of repeatability regarding US acquisition, which negatively impacts the implementation of extensive screening programs.}

\par
\revision{In contrast to conventional free-hand US, automatic screening systems, with \revision{stable} acquisition quality, can enable regular checks for PVD.}
\revision{Automating the screening procedure by employing a robotic arm can reduce personnel costs and allow sonographers to focus more on analyzing US scans.}
\revision{Furthermore, sonographers can benefit from such a system, as it reduces work-related musculoskeletal disorders.}


\par
To develop fully or semi-autonomous robotic US systems (RUSS) and further obtain enhanced precision \revision{in} US acquisition, stable and reproducible robotic systems have been widely used in recent \revision{studies}. Pierrot~\emph{et al.} developed a RUSS able to apply a constant force between \revision{the} probe and contact surface~\cite{pierrot1999hippocrate}. Gilbertson~\emph{et al.} designed a \revision{one degree of freedom (DoF)} RUSS \revision{that reduces} image variations by providing a constant contact force~\cite{gilbertson2015force}. Conti~\emph{et al.} proposed a collaborative RUSS \revision{that provides} an interface for 3D volumetric representation obtained with constant force~\cite{conti2014interface}. \revision{In addition to} contact force, Huang~\emph{et al.} adjusted the probe orientation based on a depth camera~\cite{huang2018robotic}. To accurately position the probe in the normal direction of \revision{an} unknown object, Jiang~\emph{et al.} utilized real-time US images and force data for a convex probe~\cite{jiang2020automatic}. Then, they further developed a mechanical force model based method to accurately position different types \revision{of} probes for repeatable US acquisition~\cite{jiang2020automatic2}. 
\revision{However, to estimate the normal direction, the probe has to be rotated at a given contact position, which limits the use of these approaches when the probe is moved along a certain trajectory, namely a US sweep.} In addition, to track an object, Chatelain~\emph{et al.} employed a visual servoing method to maintain the target horizontally centered in the US image using a US confidence map~\cite{chatelain2016confidence}. But the method is not time-efficient, which took around $10~s$ to reduce the centering error to less than $3~mm$.

\par
In this work, we proposed an end-to-end workflow for autonomous robotic screening of tubular structures based only on real-time US images. This manner was inspired by the way sonographers conduct US scans. \revision{First,} the US probe is roughly placed \revision{on} the target tissue. Then, the system can automatically screen along the targeted segment of blood vessels with an optimized pose and estimate the radius of the vessel in real-time. To achieve this, a neural network was trained to segment and track the tubular structures from cross-sectional US images during the screening process. 
To the best of our knowledge, this work is the first method able to automatically perform a US scan of a vascular structure with optimized probe position and orientation in real-time. The main contributions of this work can be summarized as follows:

\begin{itemize}
  \item An end-to-end workflow to automatically perform US scans of tubular structures using real-time US images.

  \item \revision{Automatic tuning of the probe orientation and position such that the probe aligns with the normal direction of the structure of interest ($3.7\pm1.6^{\circ}$ in $2.6\pm1.5~s$) and the structure is centered in the US view ($0.2\pm0.2~mm$ in $2.0\pm1.0~s$) during sweeps.} 

  \item \revision{Starting from a random initial probe orientation within $45^{\circ}$ from the optimal orientation, the radius of the tubular tissue was accurately ($0.77\pm0.4~mm$) \secondrevision{estimated} in a short time ($1.7\pm0.5~s$).}
 
\end{itemize}
Finally, the vascular segmentation method was validated on phantoms and volunteers' arms (brachial artery). The \revision{entire} screening process was validated both by simulation and on a physical gel phantom.

\par
The rest of this paper is organized as follows. Section II presents related work. The implementation details of the neural network are presented in Section III. Section IV describes the details of the proposed close-loop controller. The experimental results are provided in Section V. Finally, Section VI and Section VII are the discussion and conclusion, respectively, of the presented method. 

\section{Related Work}

\subsection{US Image Segmentation}
\par
US is one of the most challenging modalities for automatic segmentation due to acoustic shadow, poor contrast, the presence of speckle noise, \revision{and deformation}~\cite{mishra2018ultrasound}. 
To automatically segment vessels from cross-sectional US images, Guerrero~\emph{et al.} applied an extended Kalman filter and an elliptical vessel model to determine the vessel boundary using an iterative radial gradient algorithm~\cite{guerrero2007real}. However, this method requires careful selection of a seed point inside the boundary. Furthermore, tracking between subsequent frames often fails when the transducer is moved rapidly or if the vessel is deformed.
To eliminate the need for manual initialization, Smistad~\emph{et al.} proposed an automatic artery detection method and used it to initialize vessel tracking~\cite{smistad2015real}. In addition, to robustly track vessels in a more realistic case, Crimi~\emph{et al.} employed the Lucas-Kanade method to follow vessels during sudden lateral transducer movements, even when the vessel is deformed~\cite{crimi2015automatic}. Nevertheless, the employed ellipse template matching process is \secondrevision{time-consuming}, and therefore, unsuitable for real-time applications. 

\par
Besides elliptical models, adaptive active contours were developed to achieve good segmentation results for clips when variations between US images are taken into consideration~\cite{karami2018adaptive}. 
However, this technique requires manual segmentation of the first frame.
In addition, Hessian matrix based methods, \revision{such as the Frangi filter~\cite{frangi1998multiscale}, have been} developed to extract tubular structures. 
\revision{But this} method \revision{cannot} accurately extract vessel outlines for \revision{further} planning US screening.  


\begin{figure*}[ht!]
\centering
\includegraphics[width=0.90\textwidth]{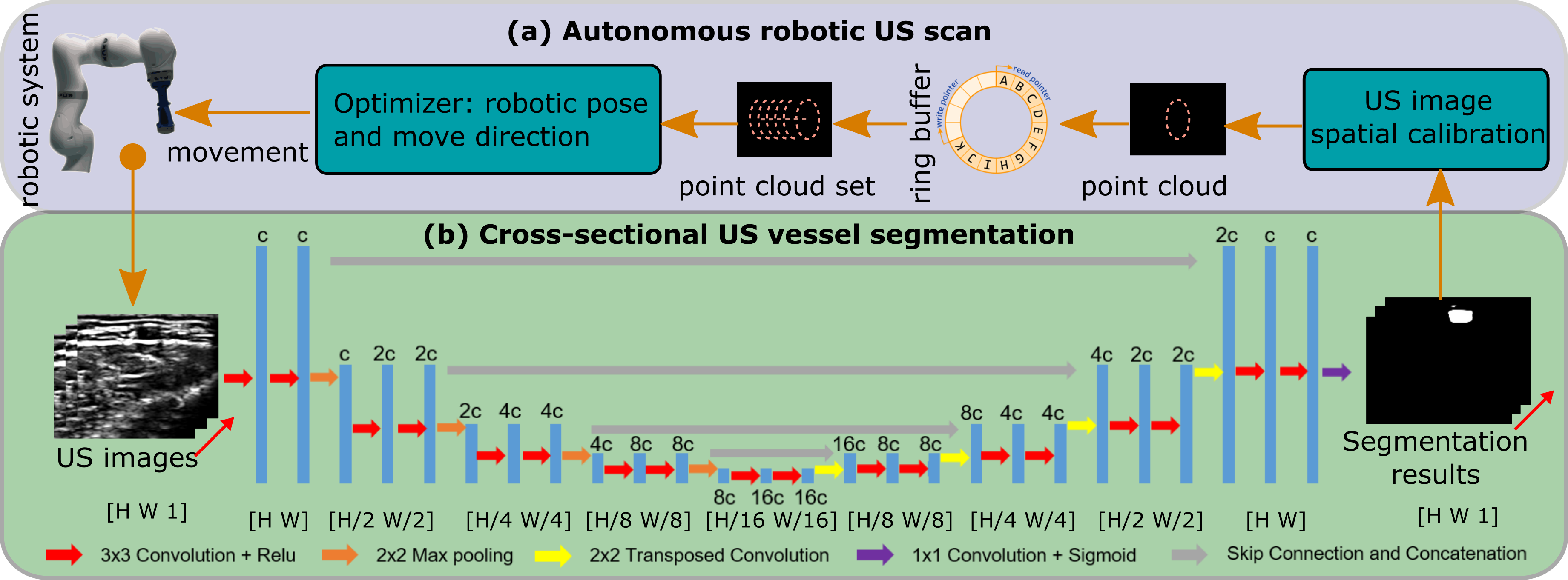}
\caption{Autonomous image-guided robotic US vessel screening. (a) automatically moving along the target vessel based on real-time segmented results. (b) extracting vessels \revision{from} cross-sectional US scans using U-Net.\revision{The letter``c" represents the number of image channels and ``H" and ``W'' represent the height and width of images, respectively.}
}
\label{Fig_general_process}
\end{figure*}

\par
\revision{Recently, deep learning has been introduced as a promising alternative to classical US segmentation algorithms.}
Convolutional neural networks (CNN) achieved phenomenal success in various computer vision tasks and \revision{have been successfully} applied to CT and MRI segmentation tasks~\cite{litjens2017survey}. However, their use in US image segmentation is limited due to acoustic shadow, poor contrast, and the presence of speckle~\cite{mishra2018ultrasound}. Ronneberger~\emph{et al.} proposed U-net architecture for biomedical image segmentation~\cite{ronneberger2015u}. The network is extended from a fully CNN and is considered \revision{an effective method} for medical image segmentation. Mishra~\emph{et al.} developed a fully CNN with attentions for automatic and accurate segmentation of US images~\cite{mishra2018ultrasound}. Chen~\emph{et al.} employed a recurrent fully CNN architecture to capture salient image features and motion signatures at multiple resolution scales~\cite{chen2020deep}. 

\subsection{Robotic US Screening}
\par
RUSS \revision{has} also been developed to introduce autonomous and accurate health care solutions for vascular diseases. Balter~\emph{et al.} developed a 7-DoF robot for venipuncture using 3D \secondrevision{near-infrared} and US imaging of peripheral blood vessels~\cite{balter2016adaptive, chen20163d}.
Langsch~\emph{et al.} proposed a \revision{US-based} method for catheter tracking and visualization during endovascular aneurysm repair \revision{while} the search path is limited to a vessel path computed from a pre-operative MRI image~\cite{langsch2019robotic}. Virga~\emph{et al.} developed an autonomous framework to acquire abdominal 3D US images for abdominal aortic aneurysms~\cite{virga2016automatic}. To automatically obtain a suitable trajectory for different patients, the patient surface obtained from an RGB-D camera was registered to a generic MRI-based atlas. 
This means the system cannot handle unexpected cases, such as \revision{when} the target vessel is not visible in the US view. In addition, the vessel radius still needs to be manually measured using the 3D US volume.


\subsection{Proposed Approach}
\par
In order to online optimize the probe orientation and position and estimate the vessel geometry in real-time, a neural network was employed to provide robust and accurate vascular segmentation results as feedback for a \secondrevision{closed-loop} controller. 
\revision{The most recent segmentation results were used to characterize the local vessel outline in 3D. Based on \revision{the} 3D data, the vessel centerline and the radius can be estimated by solving an optimization problem.}
\revision{Based on the estimated vessel centerline}, a RUSS can automatically screen blood vessels with no need for pre-operative images.
To validate the performance of the segmentation network on in-vivo human tissues, the brachial artery of the forearm was used as the target object in in-vivo tests. Besides, to ensure the safety of involved objects, a simulator with known vessel boundaries was built to theoretically validate the proposed control framework. After fully testing the controller with varied initial settings by simulation, the end-to-end workflow for automatic US scanning (Fig.~\ref{Fig_general_process}) was entirely validated on a physical gel phantom\footnote{The video: https://www.youtube.com/watch?v=VAaNZL0I5ik}.

\section{Segmentation and Tracking Algorithm}
\par
To robustly and accurately segment tubular tissue in real-time from cross-sectional US images, U-Net architecture was employed, \secondrevision{consisting} of an encoder and decoder. 
The encoder has a pyramid structure, which is commonly used for segmentation tasks. The decoder contains up-sampling layers that enable the generation of a pixel-wise segmentation mask. Skip connections between encoder and decoder provide the decoder with more image features from the shallow layers, which is beneficial for segmenting detailed structures. \revision{The structure of the U-Net is described in Fig.~\ref{Fig_general_process} (b).}


\par
The batch size was set to ten due to GPU memory constraints. This small batch size also improves the generalizability. However, such a small batch size interferes with the batch normalization layers used in the original U-Net architecture. Thus, group normalization~\cite{wu2018group} was employed in this work. 


\par

\revision{The U-Net has been proven \revision{as} a promising method to segment the vessels from US images~\cite{prevost20183d}. However, to train a generic model with good performance for both phantom and human data, a large dataset has to be prepared as the textures of phantom and human images differ. Thus, two models with the same network architecture were trained separately for phantom and human images, respectively.} 
The model developed for an in-vivo human blood vessel (brachial artery) was trained using $1,219$ US images ($256\times256$ pixel) acquired from three volunteers \revision{(BMI: $23.2\pm0.5$, age: $28\pm2$)}, and the model developed for vascular phantoms was trained based on $3,262$ US images ($256\times256$ pixel). 
\revision{In-vivo images of brachial arteries were recorded of the anterior view of the forearm starting below the elbow and towards the wrist with a total distance of $60~mm$. At least two sweeps were recorded for each subject. To ensure class balance, $1,219$ images were averagely taken from different subjects (around $400$ images for each subject).}
\revision{The phantom images were recorded from two custom-designed vascular phantoms with different vessel radii. The larger radius ($7.5~mm$) was used to mimic big vessels, like the carotid and aorta, while the smaller one ($4~mm$) was used to mimic small vessels, like the brachial artery.} \revision{To create random textures in the phantom US images, paper pulp was mixed into the melted gelatin liquid.}
\revision{
Then, multiple US sweeps were performed along the vessels with random and varying probe orientation and contact force. Finally, $3,262$ US images were recorded from the two vascular phantoms. 
}

\par
Since US images are sensitive to acquisition parameters and the optimal parameters vary by patient and phantom, it is necessary to provide a robust segmentation for images obtained using different acquisition parameters.
To address this issue, the training images were recorded using \secondrevision{a variety of} parameters (depth, focus, and brightness). \revision{Based on the preset file for arterial scans from the manufacturer}, the dynamic range and frequency were set to $88~dB$ and $7.6~MHz$, respectively. To enable generalizability of the trained models, these parameters were randomly changed in $[70,95]~dB$ and $[6.0,8.5]~MHz$, respectively, to enhance the diversity of the data sets. 
Both data sets were carefully annotated with the assistance of a clinician. 
\revision{The dice coefficient is a popular metric used to measure the similarity between the segmentation result and labeled images. Based on this metric, the loss function of the U-Net was designed as follows:
} 

\revision{
\begin{equation}\label{eq_dice_loss}
L_d = 1 - C_d
\end{equation}
where $C_d = \frac{2~|G\cap S|}{|G|+|S|}$ is the dice coefficient. In $C_d$, $G$ are the labeled images, in which the object is carefully annotated, and $S$ are the output images of the network predicting the target object position in US images. 
}



\section{Force-Compliant Robotic US Screening} \label{section-automatic-screening} 
This Section describes the method used to optimize the vessel centerline and robotic pose. The optimization was purely based on the most recent US images, without the requirement \revision{of} additional devices or pre-operative images. 

\subsection{System Calibration}

\begin{figure}[ht!]
\centering
\includegraphics[width=0.48\textwidth]{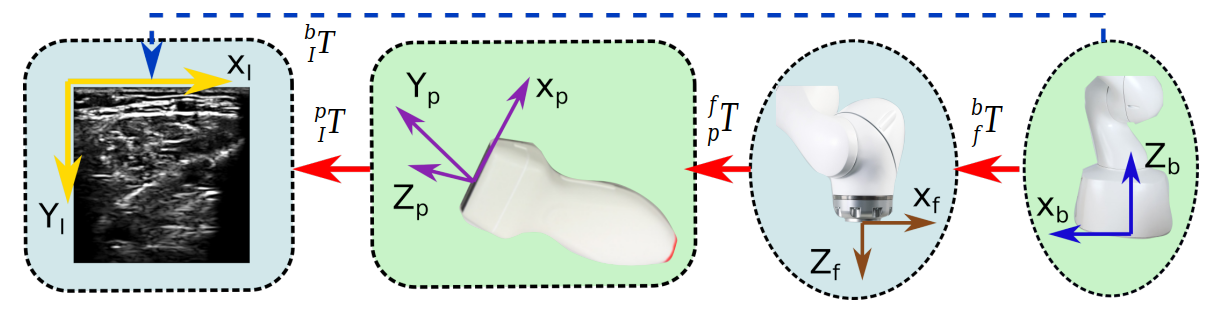}
\caption{Diagram of involved coordinate systems.
}
\label{Fig_coordinate_system}
\end{figure}

\par
In order to use the US image to control the position of the robotic manipulator, the matrix $^{b}_{I}\textbf{T}$ transforming pixel position \revision{from} a US image $\{I\}$ into the robotic base frame $\{b\}$ should be calculated. The involved frames are depicted in Fig.~\ref{Fig_coordinate_system}. Thus, $^{b}_{I}\textbf{T}$ can be calculated as follows:
\begin{equation}\label{eq_trasformation}
^{b}_{I}\textbf{T} = ^{b}_{f}\textbf{T}~^{f}_{p}\textbf{T}~^{p}_{I}\textbf{T}
\end{equation}
where $^{j}_{i}\textbf{T}$ is the transformation matrix used to transfer the position from frame $\{i\}$ to frame $\{j\}$. Frames $\{p\}$ and $\{f\}$ are the probe tip and robot flange frames, respectively. 

\par 
Since the kinematic model of the robot was given, the transformation $^{b}_{f}\textbf{T}$ from robotic flange to robotic base frame can be obtained using the software API provided by the manufacturer. As for $^{f}_{p}\textbf{T}$, it will be fixed once the probe is attached to the flange. To simplify the calculation and avoid error caused by data truncation, the probe was installed parallel to \revision{the} frame $\{f\}$ ($^{b}_{f}\textbf{R}=(1,0,0; 0,1,0; 0,0,1)^T$ or $(-1,0,0; 0,-1,0; 0,0,1)^T$). In addition, the frame origin was set to the central point of \revision{the} probe tip (Fig.~\ref{Fig_coordinate_system}). 
The translational part of $^{f}_{p}\textbf{T}$ was obtained from the 3D model. 

\par
As for the linear probe, the US elements are physically distributed on the tip within a specific length $L_p$, which can be obtained from the specification provided by the manufacturer. \revision{Based on the given imaging depth $D_I$ and $L_p$, the pixel position $(u, v)$ can be mapped in physical length $(^{I}x,^{I}y) = (\frac{L_p}{\secondrevision{H}}u, \frac{D_I}{\secondrevision{W}}v)$, where $\secondrevision{H}$ and $\secondrevision{W}$ are the height and width (in terms of pixel) of the B-mode images, respectively. The origin of frame $\{I\}$ was set at the upper left of the B-mode image, while the origin of frame $\{p\}$ was set in the middle of the probe tip (Fig.~\ref{Fig_coordinate_system}).} 
Thus, $^{p}_{I}\textbf{T}$ is calculated as follows:

\begin{equation}\label{eq_I_to_probe}
^{p}_{I}\textbf{T} = \begin{bmatrix}
\frac{L_p}{\secondrevision{H}} & 0 & 0 & -\frac{L_p}{2}\\
0 & 0 & -1 & 0\\
0 & \frac{D_I}{\secondrevision{W}} & 0 & \varepsilon_0\\
0 & 0 & 0 & 1\\
\end{bmatrix}
\end{equation}
where $L_p=37.5~mm$, $D_I = 40~mm$, and $\varepsilon_0$ is used to represent a small distance from probe frame origin to image frame origin determined by the US element configuration.   

\subsection{Impedance Control Architecture}
\par
Safety is the most important issue for the development of automatic robotic systems, especially for medical robots. In order to avoid any damage to patients, an impedance controller \revision{using built-in joint torque sensors} was employed~\cite{hennersperger2016towards}. The involved Cartesian compliant control law has been defined as

\revision{
\begin{equation}\label{eq_impedance_law}
\tau = \textbf{J}^{T}[\textbf{F}_d + \textbf{K}_m e + \textbf{D} \dot{e} + \textbf{M} \Ddot{e}]
\end{equation}
}
where $\tau \in \textbf{R}^{7\times1}$ is the computed target torque for all joints, \revision{$J^{T}\in \textbf{R}^{7\times6}$} is the transposed Jacobian matrix, $e\in \textbf{R}^{6\times1} = (x_d - x_c)$ is the pose error (\secondrevision{position and orientation}) between the current pose $x_c$ and the target pose $x_d$ \secondrevision{in Cartesian space}, $\textbf{F}_d\in \textbf{R}^{6\times1}$ is the supposed exerted force/torque at endeffector, \secondrevision{$\textbf{K}_m \in \textbf{R}^{6\times6}$, $\textbf{D}\in \textbf{R}^{6\times6}$ and $\textbf{M}\in \textbf{R}^{6\times6}$ are diagonal matrices of stiffness, damping and inertia terms in 6 DoF, respectively.}

\par
The impedance controller works like a spring with a given stiffness $\textbf{K}_m$. If an unexpected obstacle is present between the current position and the target position, such as a certain body part of the patient, the robot will stop at the contact position with a certain force. However, if the resistance force is not sufficiently large, the manipulator will overcome the resistance force and continue moving towards \revision{the} target position.
According to~\cite{hennersperger2016towards}, stiffness in the direction of the probe centerline is usually set in the range $[125, 500]~N/m$ for human tissues.
Thus, $\textbf{K}_m$ was set to \thirdrevision{diag(}$1000~N/m, 1000~N/m, 300~N/m, 2~Nm/rad$, $20~Nm/rad, 20~Nm/rad$\thirdrevision{)} representing \revision{the translational and rotational stiffness in X, Y, and Z and Z, Y. and X direction, respectively. The stiffness in the X and Y directions were empirically set to $1000~N/m$ to balance the positioning accuracy and compliant propriety in these directions and avoid hard collision.
} 
\secondrevision{In addition, the damping ratio in all 6 DoFs was set to $0.8$.} 
To determine a safe force for patients, \revision{the target force $F_d$ of the controller was gradually increased on a volunteer's arm.} Finally, $25~N$ was set as a software restriction to limit the force exerted by the robot. If \secondrevision{contact force} $F_c>25~N$, the robot will automatically stop to avoid an excessive force.

\subsection{Estimation of the Vascular Structure Centerline}
\par
To build a closed loop system, the robotic movements were optimized based on the received US images as shown in Fig.~\ref{Fig_general_process}. To make the movements smooth and robust, a ring buffer was used to save the newest $N_R$ point clouds generated by transforming the segmentation results from the neural network into the coordinate frame of the robot. Since the cross-section of human blood vessels can be \secondrevision{modeled} as ellipses~\cite{guerrero2007real,smistad2015real,crimi2015automatic}, the extracted point clouds depicting the vessel boundary can be seen as an elliptic cylindrical surface. Then the \revision{vector} of vessel centerline  ($\Vec{n}_v = (n_1, n_2, 1)$) and the vessel \revision{radius} ($r_v$) are estimated by solving the following optimization problem:


\begin{equation}\label{eq_longititude_direction_cost}
\begin{split}
&\min_{\Vec{n}_v, r_v, \varepsilon}
\underbrace{
\frac{1}{2N}\sum_{i=1}^{N}\left((\frac{\|\overrightarrow{CP_i}\times\Vec{n}_v\|}{\|\Vec{n}_v\|}-r_v)^2+\varepsilon^2\right)}_{\text{optimization term}} \\
&~+ 
\underbrace{
\frac{\lambda_1}{2}\left(\arctan{\frac{n_2}{n_1}}-\arctan{\frac{n_2^{'}}{n_1^{'}}}\right)^2}_{\text{orientation stabilization term}} + 
\underbrace{
\frac{\lambda_2}{2}(r_v-r_v^{'})^2}_{\text{radius stabilization term}}\\
&~~~~~~~~~~~~~s.t.~\left\{
\begin{split}
\varepsilon &> 0 \\
r_v &> r_l \\
r_v &\leq \varepsilon + r_h \\
\end{split}
\right.
\end{split}
\end{equation}
where $N$ is the number of points in the point cloud set, $\textbf{P}_i=(x_i, y_i, z_i)$ is the point located on the detected vessel boundary,  $\textbf{C}=\frac{1}{N}\sum_{i=1}^{N}\textbf{P}_i$ is the centroid of the point cloud set. The cross product item represents the distance from $\textbf{P}_i$ to the estimated centerline $\Vec{n}_v$.
$r_l$ and $r_h$ are the lower and higher bound of the vessel radius, $r_v\in~(r_l, r_h)$, \secondrevision{$\varepsilon$} is used to soften the hard constraint of the higher bound, to adapt to a serious arterial aneurysm case, wherein the vessel \revision{radius} becomes much larger than normal. \revision{$n_1^{'}$, $n_2^{'}$} and $r_v^{'}$ represent the last optimized results. The latter two terms are used to stabilize the motion of the manipulator. \revision{In order to minimize the loss function (Eq.~\ref{eq_longititude_direction_cost}), a large pose or radius deviation from the current pose or radius has to be avoided during the optimization process.}
$\lambda_1$ and $\lambda_2$ are \revision{hyper parameters for tuning the performance of the stabilization terms. Both were set to one in this work.}

\begin{figure}[ht!]
\centering
\includegraphics[width=0.40\textwidth]{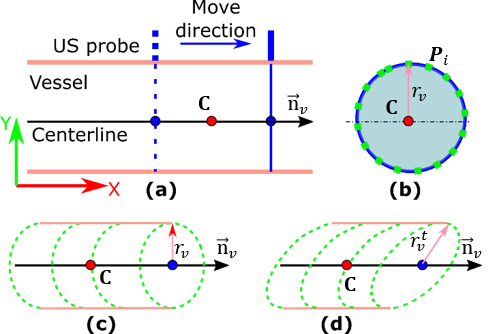}
\caption{(a) Front and (b) cross-sectional views of blood vessel acquired when the probe is placed normal to the blood vessel. \revision{(c) and (d) are the detected point clouds for the sweep when the probe was placed in normal direction and titled direction of vessel centerline, respectively.} 
}
\label{Fig_ComputeRadius}
\end{figure}

\revision{Considering that the vessel geometry in 2D images is affected by the probe orientation, if the probe is titled form the normal direction, the radius in the 2D image $r_v^t$ (Fig.~\ref{Fig_ComputeRadius} (c)) will be larger than the real $r_v$ (Fig.~\ref{Fig_ComputeRadius} (d)). To accurately estimate $r_v$ in all cases, the newest $N_R$ 3D point clouds are saved in a ring buffer and used to characterize vessel geometry. Then $r_v$ can be approximated by $\frac{1}{N}\sum_{i=1}^{N}\frac{\|\overrightarrow{CP_i}\times\Vec{n}_v\|}{\|\Vec{n}_v\|}$.}

\par
The constrained non-linear optimization problem Eq.~(\ref{eq_longititude_direction_cost}) was solved by implementing a sequential quadratic programming (SQP) optimizer using \revision{the} NLopt library. Due to the cross product operation, the gradients of the objective function with respect to $n_1$ and $n_2$ \revision{were} computed using the Symbolic Math Toolbox in MATLAB R2020 (MathWorks, USA).
However, such an optimizer can get stuck in a local minimum. In our case, the optimizer sometimes yielded a resulting direction vector pointing in the radial direction of the cylinder. In order to circumvent this issue, the original optimization problem was implemented in an iterative \revision{``}Tick-Tock" manner. 
The radius $r_v$ was first fixed to a given value, and only the direction vector $\Vec{n}_v$ was optimized. Then, $r_v$ was optimized separately with a fixed direction vector. These two steps are executed once in each iteration so that the result will converge to an acceptable local minimum, or even to the global minimum. 

\par
Since the \revision{frame rate} of the US images is high ($50~fps$), the detected point clouds in the ring buffer may heavily overlap.
This will hinder the optimization process, particularly for real cases wherein the vascular segmentation results are not as perfect as those in a simulation.
\revision{Considering an extreme case when the probe is paused at a certain position, the detected vessel boundary point clouds in the ring buffer will be distributed on a plane. In such a case, the optimization process could be stuck at a local optimal solution of $\Vec{n}_v$ at $\overrightarrow{CP_i}$ (Fig.~\ref{Fig_ComputeRadius}).}
Reducing the overlap will help the optimizer converge to the correct solution (\secondrevision{centerline} line $\Vec{n}_v$ and estimated mean \revision{radius} $r_v$ of the vessel).
\revision{To avoid sub-optimal results}, the distance between the point clouds $\Omega_j$ and $\Omega_1$ in the ring buffer were scaled as follows:


\begin{equation}\label{eq_enlarge_distance}
\textbf{P}_i^{'} = \textbf{P}_i + \mu(\textbf{C}_j-\textbf{C}_1)~~~~~~~~\forall~P_i \in \Omega_j
\end{equation}
where $\textbf{P}_i$ is a point in $j$-th point cloud $\Omega_j$, $\textbf{C}_j$ and $\textbf{C}_1$ are the centroids of the $j$-th and the first point cloud, respectively, $\mu$ is a constant coefficient to move the two point clouds away from each other along the line connecting two centroids. \revision{$\mu = 5$ in this work}.

\subsection{Determination of Robotic Pose}
\par 
In order to obtain a high-quality US image, the US probe is usually aligned in the normal direction of the examined object to \revision{reduce the number of scattered echoes}~\cite{jiang2020automatic, huang2018robotic, ihnatsenka2010ultrasound}.
However, previous work~\cite{jiang2020automatic, huang2018robotic} only considered the normal direction of the skin surface above target tissues. 
In order to find the normal direction of a vessel tissue, which is located below the skin surface, both camera-based~\cite{huang2018robotic} and force-based methods~~\cite{jiang2020automatic, jiang2020automatic2} are not applicable.


\par
However, as detailed in the previous Section, the present method can estimate the centerline vector $\Vec{n}_v$ of a blood vessel from a set of US images; allowing normal alignment of the US probe to the target vessel by aligning the probe's $Y_p$ axis with $\Vec{n}_v$.
Once the alignment has been done, the US image will remain normal to the vessel when the probe is rotated around $\Vec{n}_v$. However, this rotation will result in non-homogeneous deformation in both sides of the US image or even cause a gap between \revision{the} US probe and skin. Thus to fully obtain the proper probe orientation, and to achieve the best US image quality, the probe $Z_p$ axis is aligned with the normal of the contact surface $\Vec{n}_s$. Since this is evaluated on a gel phantom with a flat surface, $\Vec{n}_s$ is approximated using the normal direction of a plane composed of three neighboring points manually selected using the robot. After this procedure is completed, the target probe orientation is determined ($X_p = Y_p\times Z_p$).


\par
In addition, to maintain the target vessel horizontally centered in the US image for a better view, the position of the probe also needs to be adjusted based on US images. 
Since the relative movement $\Delta P$ is continuously updated using real-time US images, the target vessel is able to be maintained around the horizontal center of the US view, even when the displayed tissue is deformed due to the contact of the probe.
\begin{equation}\label{eq_centralization}
\Delta P =~^{b}_{I}\textbf{T}~\left[\frac{H}{2}-x_c^{I},~0,~0,~1\right]^{T}
\end{equation}
where $x_c^{I} = \frac{1}{N}\sum_{i=1}^{N}x_p^i$ is the horizontal center of the detected vessel in the current image, and $x_p^i$ is the \revision{horizontal value of the point $\textbf{P}$ in the ring buffer.}


\section{Results}

\subsection{Experimental Setup}
\begin{figure*}[ht!]
\centering
\includegraphics[width=0.71\textwidth]{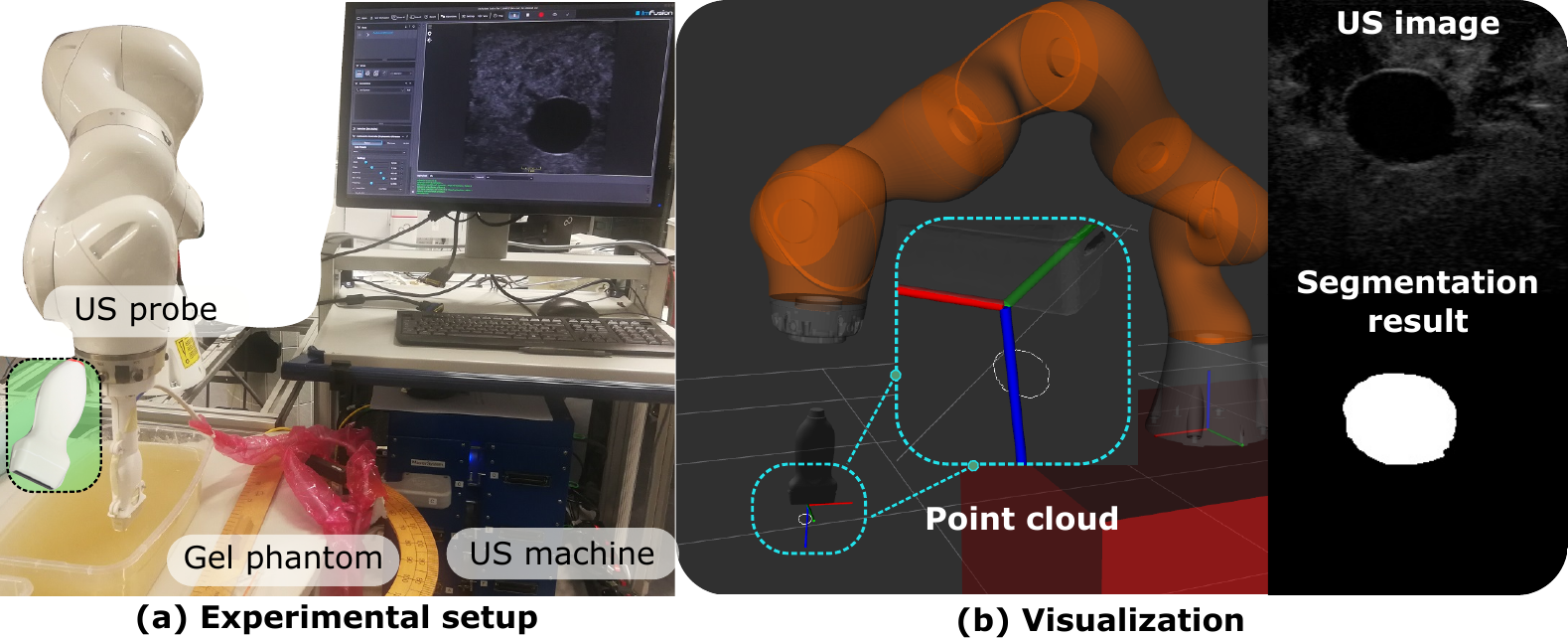}
\caption{(a) Physical setup of experiments on a gel phantom. (b) Real-time visualization in RVIZ. The real-time US image and segmentation results were also visualized in RVIZ.}
\label{Fig_experimentsetup}
\end{figure*}

The overall experimental setup is shown in Fig.~\ref{Fig_experimentsetup}~(a). A linear probe (CPLA12875, Cephasonics, California, USA) was attached to the end-effector of a robotic manipulator (KUKA LBR iiwa 14 R820, KUKA Roboter GmbH, Augsburg, Germany) using a 3D-printed mount. \revision{The used robot has accurate built-in torque sensors in all joints.}
The B-mode US images ($50$ fps) \revision{were} obtained using a Cephasonics machine (Cephasonics, California, USA) via USB interface and the images \revision{were} visualized using ImFusion software (ImFusion GmbH, Munich, Germany). To access real-time images from different computers, the US image stream \revision{was} published to \revision{Robot Operating System (ROS)}. In addition, the robot \revision{was} controlled via iiwa stack developed based on the ROS~\cite{hennersperger2016towards}. 
\revision{The control commands and the robot status \revision{were} exchanged at $100~Hz$.} 

\par
To build a ballistic gel phantom with non-homogeneous acoustic impedance mimicking human tissues, paper pulp ($3-5~g/L$) and \revision{gelatin powder ($175~g/L$)} were mixed into water. \revision{After solidification, two round tubes were used to create holes inside the phantom mimicking vascular structures.}
\revision{Since the radii of the tubes are constant along the centerline, the inner radius of the created holes can also be seen as a constant value, which was approximated using the average radius directly measured from the removed cylinder portion of the phantom.} 
\revision{To completely validate the proposed end-to-end automatic US scan approach, one of the phantoms, with a radius of $7.5~mm$, was used for physical experiments.}
\revision{The marching velocities for both simulation and phantom experiments in the direction $\Vec{n}_v$ were set to $1~cm/s$.}
To monitor the screening process, the RUSS and the point cloud of the detected vessel \revision{were} visualized in ros-visualization (RVIZ) as shown in Fig.~\ref{Fig_experimentsetup}.


\subsection{Segmentation Results on Phantom and Human Arm}
\begin{figure}[ht!]
\centering
\includegraphics[width=0.40\textwidth]{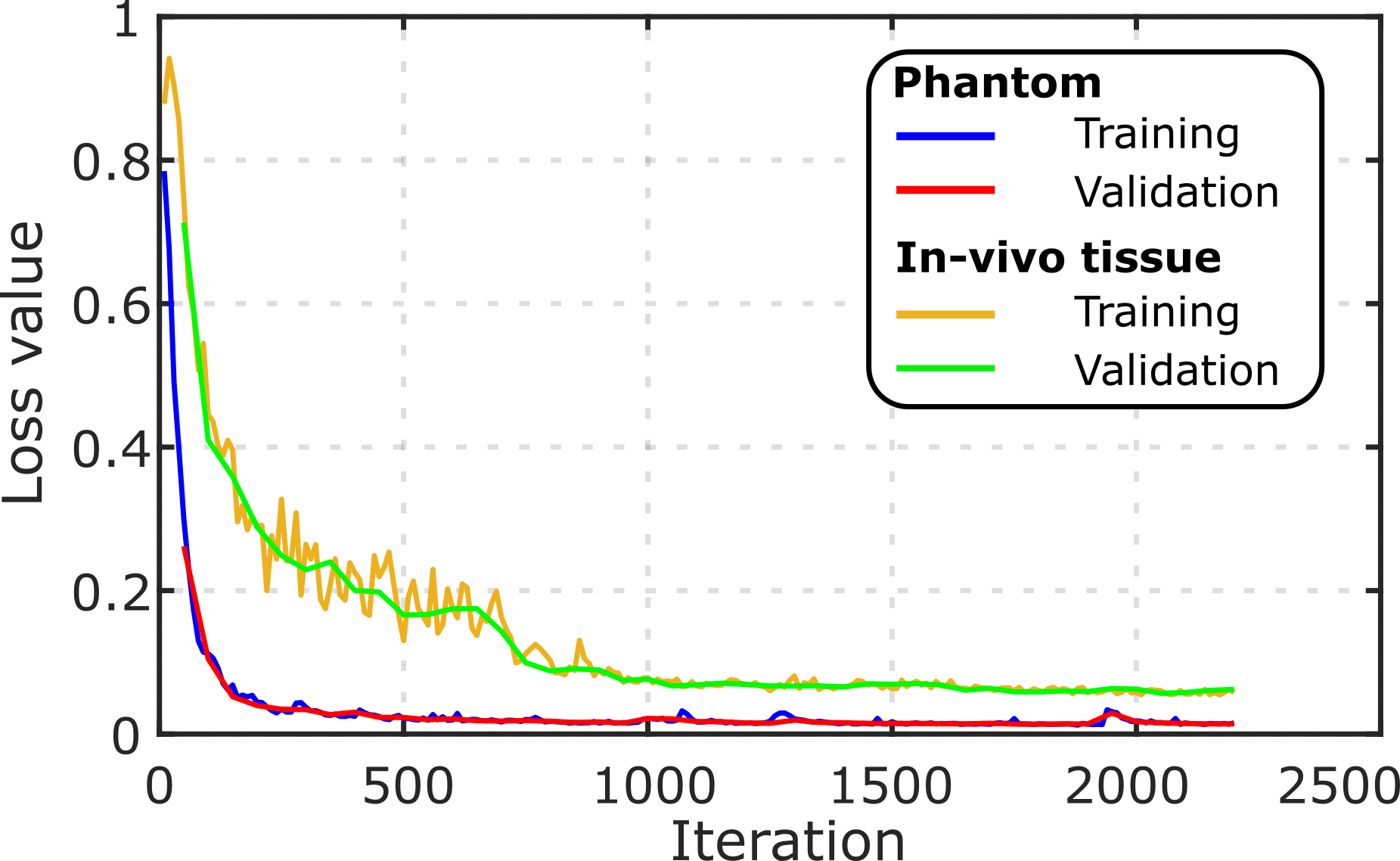}
\caption{\revision{Training loss and validation loss of the segmentation network.}}
\label{Fig_loss}
\end{figure}

The model was optimized using the ADAM optimizer~\cite{kingma2014adam} on a single GPU (Nvidia GeForce GTX 1080). The learning rate \revision{was} set to $0.001$ in the beginning and decreased by a factor of ten when the loss changes were less than $0.0001$ for ten subsequent steps.
\revision{The decrease of the learning rate in the latest step of training can help further reduce loss.}
\revision{Besides, the U-Net models were trained on $90\%$ of the images of each data set while the remaining $10\%$ of the images were used for validation.}
\revision{It can be seen from Fig.~\ref{Fig_loss} that both training loss and validation loss were quickly reduced at the beginning and gradually converged after $1,000$ iterations.}
The performance of the trained model on unseen images of brachial arteries and phantoms are summarized in TABLE~\ref{Table_segmentation_result}.

\begin{table}[!ht]
\centering
\caption{Performance of Segmentation Algorithm}
\label{Table_segmentation_result}
\begin{tabular}{cccc}
\noalign{\hrule height 1.2pt}
Dataset   & Dice Coefficient & Time (ms) \\ 
\noalign{\hrule height 1.0 pt}
Phantom   & $0.98\pm0.02$ & $5.1$\\

Brachial artery  & $ 0.93\pm0.05 $ & $5.1$\\

\noalign{\hrule height 1.2 pt}
\end{tabular}
\end{table}

\par
\begin{figure}[ht!]
\centering
\includegraphics[width=0.37\textwidth]{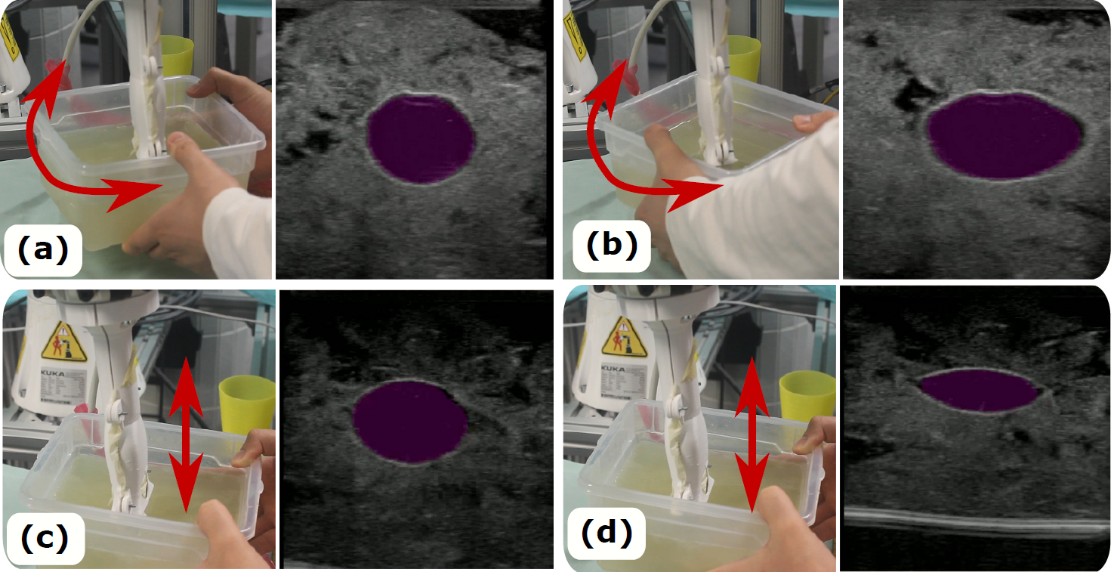}
\caption{Sample segmentation results on a phantom.
}
\label{Fig_segmentation_phantom_result}
\end{figure}
\begin{figure}[ht!]
\centering
\includegraphics[width=0.37\textwidth]{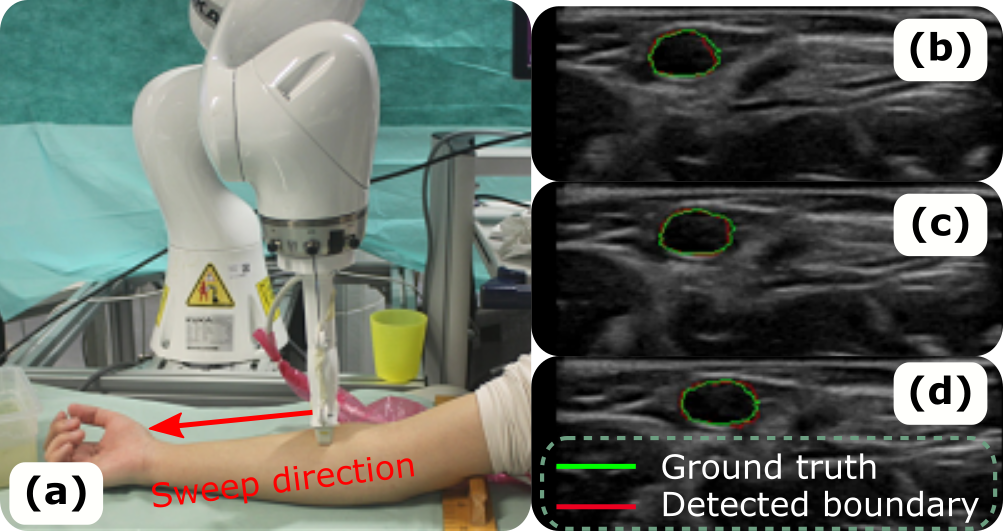}
\caption{Sample segmentation results on a human.
}
\label{Fig_segmentation_human_result}
\end{figure}

The trained model \revision{can} segment the vascular structure in $5.1~ms$ and the average dice coefficients \revision{($\pm SD$) were $0.93\pm0.05$ and $0.98\pm0.02$ for human tissue ($26$ images) and phantom ($100$ images)}.
To validate whether the trained model is able to robustly and accurately segment the target from cross-sectional US images, the phantom was rotated to different orientations relative to the probe as shown in Fig.~\ref{Fig_segmentation_phantom_result}~(a) and (b). In addition, deformation is unavoidable for US images due to a non-zero contact force. \revision{To validate the performance of the proposed segmentation method on deformed US images, the pressure between the probe and phantom was manually increased until the vessel was severely deformed as shown in Fig.~\ref{Fig_segmentation_phantom_result}~(d).} 
As can be seen in Fig.~\ref{Fig_segmentation_phantom_result}, the trained model is able to accurately and robustly segment the target tissue (mean dice coefficient is over $0.98$) from the US images with different extents of deformation.
In order to further consider the potential for real clinical use, we tested the performance on arms of volunteers (brachial artery). The network also successfully extracts the target vessel from a continuous sweep in the testing data set as shown in Fig.~\ref{Fig_segmentation_human_result}. The mean dice coefficient for in-vivo test was \revision{ $0.93\pm0.05$}. This means that the segmentation method \revision{can be} used on \revision{in-vivo brachial artery}.


\subsection{Automatic Screening in Simulation}
\par
\begin{figure}[ht!]
\centering
\includegraphics[width=0.37\textwidth]{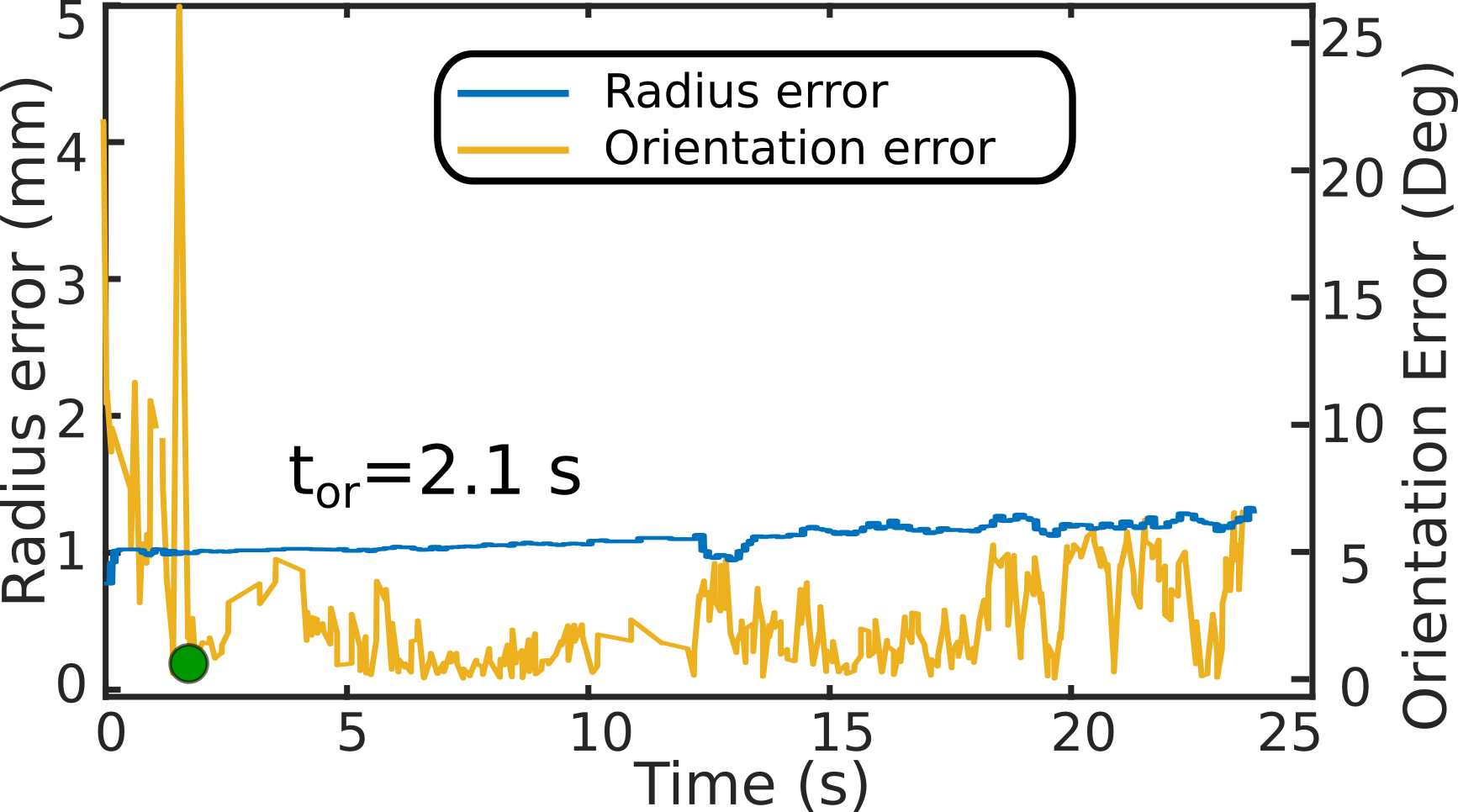}
\caption{Performance of the optimization method by simulation.
}
\label{Fig_simulation_validation}
\end{figure}

To \revision{avoid damage of objects under evaluation,} the proposed \revision{end-to-end control framework} for automatic vessel screening was first validated in a simulator. Here, the elliptical vessel boundary point clouds \revision{were} generated in MATLAB R2020 using a 2D plane to cut a cylindrical surface mesh. The resulting 2D point clouds were fed into the ring buffer to optimize the vessel radius and the vessel centerline by solving Eq.~(\ref{eq_longititude_direction_cost}). 
Then, the probe could be gradually aligned with the normal direction of the vessel and moved along the vessel centerline. 
To demonstrate the performance of the proposed approach \revision{by} simulation, the absolute \revision{radius} error $e_{ra}$ between \revision{the} estimated radius and the preset value, and the absolute orientation error $e_{or}$ between computed and desired poses are shown in Fig.~\ref{Fig_simulation_validation}. \secondrevision{Since the radius of some healthy arteries like the aorta can be $16.8~mm$ for males~\cite{mao2008normal}, the radius was set to $10~mm$ in the simulation.
}

It can be seen from Fig.~\ref{Fig_simulation_validation} that the proposed optimization method effectively decreases $e_{or}$ to $5^{\circ}$ from a random initial position in a short amount of time ($2.07~s$). The largest angular error ($22.5^{\circ}$) only occurs at the beginning of the process. 
This is because heavy overlap of the first few US images results in failure of the optimization method to find a good estimation of the vessel centerline.
Once a good estimation of the vessel centerline is found, $e_{or}$ kept within a small value. The absolute mean ($\pm$ SD) $e_{or}$ over all $330$ scans is $2.7\pm3.3^{\circ}$. 
This error is very close to the previous state-of-the-art approach for accurate positioning of US probes on a gel phantom $2.9\pm1.6^{\circ}$~\cite{jiang2020automatic2}. More importantly, the proposed method can online optimize the probe pose during sweeps while previous methods with similar accuracy only work at the given position~\cite{jiang2020automatic, jiang2020automatic2}.

\par
In addition, the estimated radius quickly converges to a stable value that is close to the given radius of $10~mm$ very fast showing that the proposed method can effectively estimate the real radius without the strict requirement to align the probe normal to the blood vessel. The absolute mean ($\pm$ SD) $e_{ra}$ is $1.16\pm0.1~mm$.
Compared with the real radius ($10~mm$) preset in the simulation, the deviation of $e_{ra}$ $0.1~mm$ is quite small ($1\%$). Therefore, we can conclude that the proposed method demonstrates the ability to stably predict the radius with an error \secondrevision{of} less than $11.6\%$.

\begin{figure*}[ht!]
\centering
\includegraphics[width=0.85\textwidth]{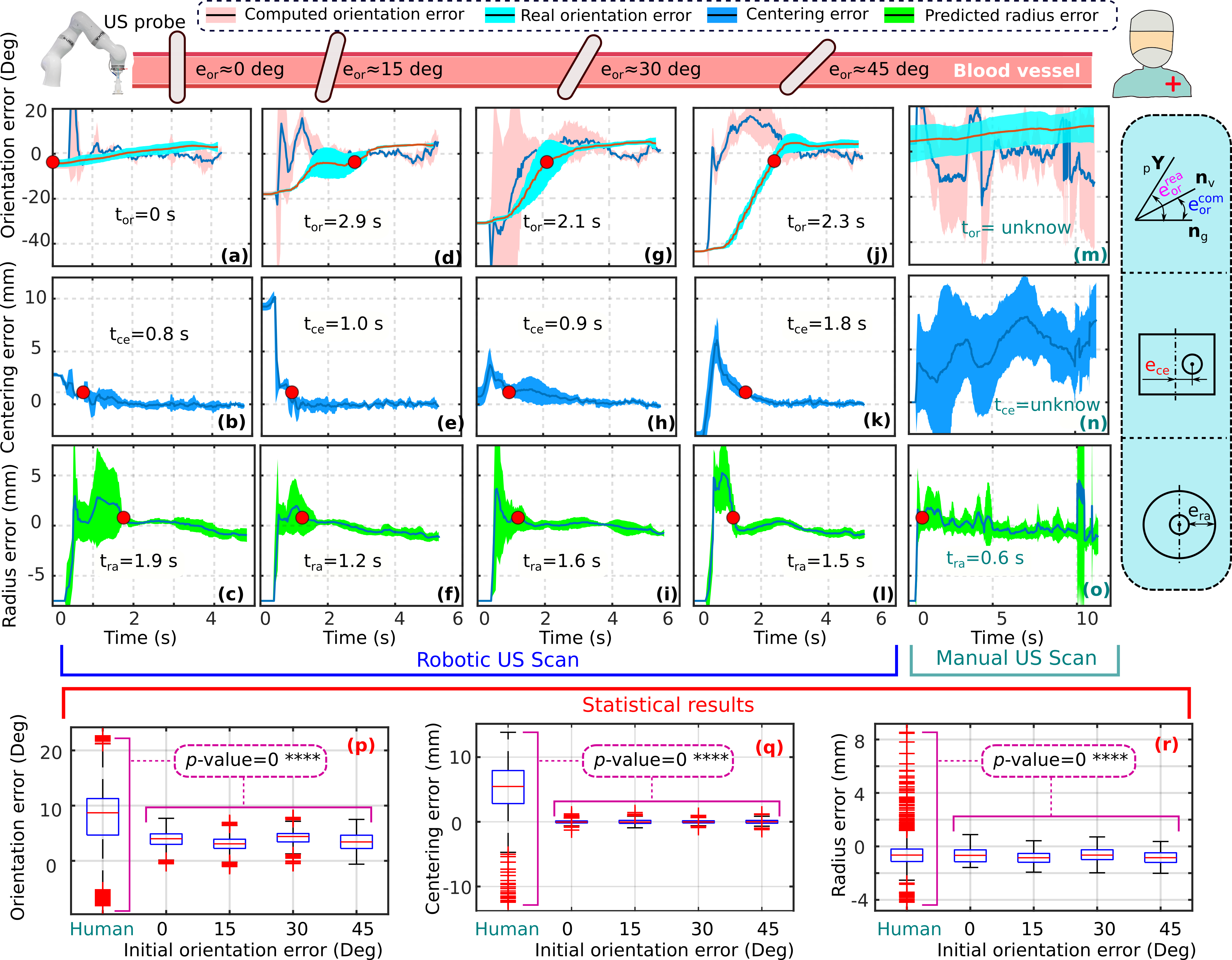}
\caption{Performance of the proposed automatic and manual screening methods on a gel phantom. The $e_{or}^{com}$, $e_{or}^{rea}$, $e_{ce}$ and $e_{ra}$ are intuitively described in the right of the figure and these errors are indicated by the peach, aqua, blue and green plots, respectively. \thirdrevision{The shadowed area represents $(mean-SD, mean+SD)$ over the repeated experiments, while the curves inside the shadowed areas are the average results.}
The results for robotic autonomous scans when the probe was placed in different initial orientations are shown in (a)-(i). (m) and (n) depict the $e_{or}^{rea}$ and $e_{ce}$ by blue and red lines, respectively. Based on US sweeps performed by a human operator, the $e_{or}^{com}$ and $e_{ra}$ computed using the proposed optimization method are described in pink and black lines in (m) and (o). (p), (q) and (r) are the statistical results of $e_{or}^{rea}$, $e_{ce}$ and $e_{ra}$ obtained from \secondrevision{the steady periods of} all experiments. \secondrevision{The $p$-value is the probability from $t$-test used to compare the performance of human operators and robotic screening in terms of $e_{or}^{rea}$, $e_{ce}$ and $e_{ra}$, respectively.} The scan path was about 8~cm.}
\label{Fig_tii_phantom_result}
\end{figure*}

\begin{figure*}[ht!]
\centering
\includegraphics[width=0.7\textwidth]{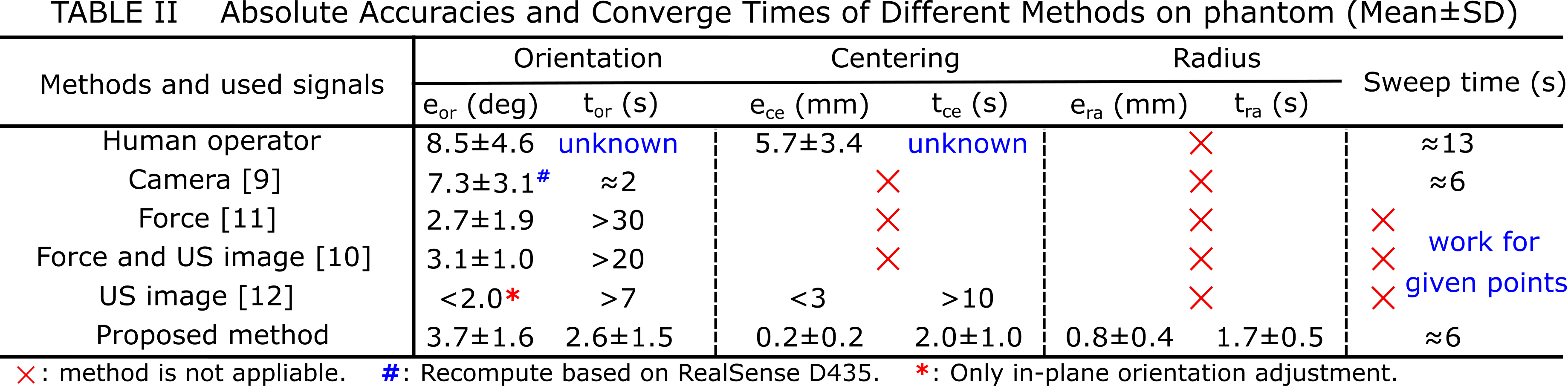}
\label{Fig_tii_method_comparision}
\end{figure*}

\subsection{Automatic Screening on \revision{a} Phantom}
\par
To further validate the performance of the proposed automatic screening system, experiments were carried out on a physical phantom with different initial orientation errors ($e_{or} \approx 0^{\circ}, 15^{\circ}, 30^{\circ}, 45^{\circ}$). 
The screening procedure was repeated ten times for each initial $e_{or}$ setting. The corresponding results \thirdrevision{($mean\pm SD$) of the robotic scans and ten manual sweeps results} on the same phantom are shown in Fig.~\ref{Fig_tii_phantom_result}. Unlike the simulation, for the real case, we calculated both the real orientation error $e_{or}^{rea}$ between the ground truth of the centerline \secondrevision{$\textbf{n}_g$} and the current robotic pose ($Y$ axis of frame $\{p\}$) as well as the computed orientation error $e_{or}^{com}$ between the computed centerline $\textbf{n}_v$ and \secondrevision{$\textbf{n}_g$} (\secondrevision{Fig.~\ref{Fig_tii_phantom_result}}). \secondrevision{To determine $\textbf{n}_g$, the probe was manually placed at the start and end points of the tubular tissue.} Then, the pixel position of the vessel centroid was transferred to frame $\{b\}$ using the spatial calibration described in Section IV-A. Thus, the vessel centerline in frame $\{b\}$ can be represented by the line connecting these two centroids. \revision{In addition, the horizontal centering error $e_{ce}$ was represented by the horizontal distance from the computed vessel's centroid of the current image and the vertical centerline of the US view.}

\subsubsection{\secondrevision{Performance of Robotic Screening}}
\par
It can be seen in Fig.~\ref{Fig_tii_phantom_result} that $e_{or}^{com}$, $e_{or}^{rea}$, $e_{ra}$ and $e_{ce}$ for all experiments with different initial $e_{or}$ can converge to zero using the proposed method. As the initial error increases from zero to $45^{\circ}$, the time $t_{or}$ required to align the probe in the desired direction also increases from zero to $2.9~s$. In most cases, $t_{or}$ is longer than the time needed to stably estimate the vessel radius $t_{ra}$. This is because the radius estimator can predict the current radius even when the probe is not placed in an optimal orientation (i.e. normal to the vessel). 
\revision{Since the robot starts moving only when the ring buffer is full, the beginning images are highly overlapped. The overlapped images will result in suboptimal results of $\Vec{n}_v$ and $e_{ra}$ as described in Section IV-C, which leads to large $e_{ra}$ and $e_{or}$ at the beginning of the optimization.}
Then, $e_{ra}$ decreases quickly to a low value near zero ($<1.8~s$) for all robotic scans. Regarding the centering error $e_{ce}$, this value approaches zero ($<0.5~mm$) after $t_{ce}$, which is the consumed time for horizontally centering the target vessel in the US view. Such small error is because the relative movement $\Delta P$ (Eq.~(\ref{eq_centralization})) can be accurately calculated with respect to pixels. 

\par
\revision{
To compare the performance of the proposed method with existing approaches, the absolute $e_{or}$, $e_{ce}$ and $e_{ra}$ and the times $t_{or}$, $t_{ce}$ and $t_{ra}$ are shown in TABLE II. The mean $e_{or}$, $e_{ce}$ and $e_{ra}$ over all $40$ experiments with different initial orientations in $[0, 45^{\circ}]$ were $3.7\pm1.6^{\circ}$, $0.24\pm0.19~mm$ and $0.77\pm0.4~mm$. Regarding probe orientation adjustment, the proposed method is not the most accurate; however, our approach would be the best if the time needed to adjust the orientation is considered ($2.6\pm1.5~s$) as well. The time needed to accurately center the object and predict the radius are $2\pm1~s$ and $1.7\pm0.5~s$, respectively. 
}

\subsubsection{\secondrevision{Performance of Manual Screening}}
\par
To compare the performance of the proposed method and of human operators, three experienced volunteers were asked to manually perform ten scans \revision{($30$ times in total)} on the same phantom. \revision{The volunteers were required to adjust the probe orientation from a random pose in the normal direction of the vessel centerline and to maintain the optimal pose during the entire sweep.}
\revision{During manual scans, the built-in ``hand-guiding" mode was activated, which allows operators to freely move the probe attached to the robotic flange. The probe pose was tracked in real-time based on the robotic kinematic model.}
One of the results was randomly selected as an example shown in Fig.~\ref{Fig_tii_phantom_result}~(m) and (n). 
Additionally, the point clouds generated from human scans were fed to the proposed optimization algorithm to predict the vessel centerline and radius. The \secondrevision{estimated} $e_{or}^{com}$ and $e_{ra}$ are depicted by pink and black lines, respectively, in Fig.~\ref{Fig_tii_phantom_result}~(m) and (o).


\par
It can be seen from Fig.~\ref{Fig_tii_phantom_result} (m) that $e_{or}^{rea}$ and $e_{ce}$ cannot converge like robotic US scans. The average $e_{or}^{rea}$ continuously increases from $3^{\circ}$ at the beginning to around $14^{\circ}$ at the end for manual scans. \thirdrevision{In addition, the deviation of $e_{or}^{rea}$ is constantly around $11^{\circ}$, which cannot be reduced as the scanning progress.}
\revision{Regarding $e_{ce}$, both the average and the deviation values cannot converge to zero. They constantly varies in the range of $[0, 8~mm]$ during the manual scan. 
This is because the human perception's ability is not good at accurately identifying small differences in orientation and position. Since such tasks require good hand-eye coordination, \revision{performing as well as RUSS} is challenging for human operators, particularly when multiple objects (optimal orientation, position, and movement along vessel) are required simultaneously.} 

\subsubsection{\secondrevision{Statistical Comparison between Robotic and Manual Screening}}
\par
\revision{To quantitatively demonstrate the advantages of the proposed approach, statistical results of $e_{or}^{rea}$, $e_{ce}$ and $e_{ra}$ of manual scans and robotic scans with different initial orientation are shown in Fig.~\ref{Fig_tii_phantom_result} (p), (q) and (r), respectively. 
For the robotic scans, the proposed method can properly adjust the probe orientation and position for centering the object in the US view, even when the initial orientation is up to $45^{\circ}$ from the surface normal.
\secondrevision{The $t$-test has been employed to compare the performance of human operators and robotic screenings in terms of $e_{or}^{rea}$, $e_{ce}$ and $e_{ra}$. The probability values $p$-value are zero, which means that there is a significant difference ($>95\%$) between the performance of manual scans and robotic scans. This phenomenon is consistent with the intuitive results witnessed from Fig.~\ref{Fig_tii_phantom_result} (p), (q) and (r) that the stable errors for robotic screenings are much smaller than the ones achieved in manual scans. }
The median $e_{or}^{rea}$ and $e_{ce}$ are $8.7^{\circ}$ versus $3.7^{\circ}$ and $3.1~mm$ versus $0.02~mm$ for manual scans and robotic scans, respectively. Since the centering movement for robotic scans was controlled in terms of pixels, $e_{ce}$ can be very close to zero. Regarding $e_{ra}$, human operators were not able to measure the vessel radius in real-time during US sweeps. Since the proposed method can estimate the radius based on the most recent frames, the method was used to predict the radius for the manual scans in Fig.~\ref{Fig_tii_phantom_result} (r). The results demonstrate that the radius prediction method can also effectively predict the radius. The median $e_{ra}$ for manual scans was $-0.7~mm$, which is very close to the robotic scans $-0.8~mm$. But it can also be observed that the maximum values of absolute $e_{or}^{rea}$, $e_{ce}$ and $e_{ra}$ for manual scans are much larger than the errors for robotic scans ($22.6^{\circ}$ versus $7.8^{\circ}$, $13~mm$ versus $1.4~mm$ and $8.5~mm$ versus $2~mm$). This is due to the manual scans' relative instability compared to the robotic scans (i.e., scan velocity and probe pose). 
}


\begin{figure*}[ht!]
\centering
\includegraphics[width=0.80\textwidth]{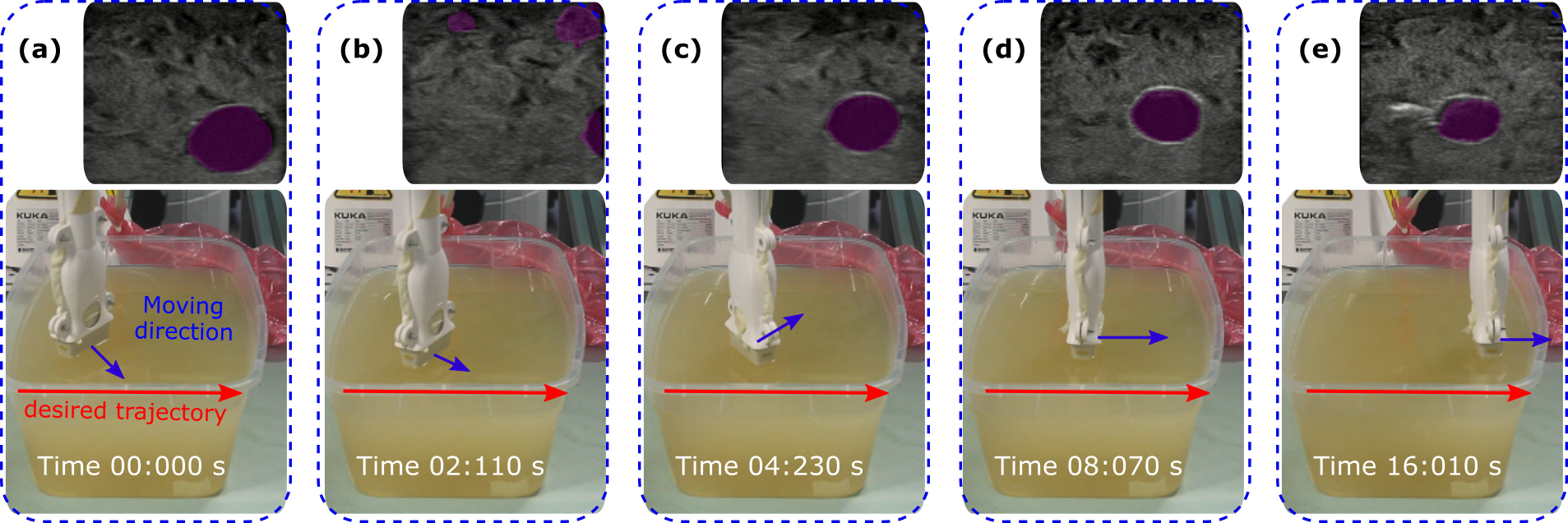}
\caption{Performance of the optimization method on a phantom with an suboptimal initialization.}
\label{Fig_phantom_result}
\end{figure*}

\subsubsection{\secondrevision{Representative Robotic Screening Procedures}}
\par
To further show the detailed process of the proposed end-to-end screening system, a full US sweep ($16~cm$) starting from a large initial $e_{or}\approx45^{\circ}$ is shown in Fig.~\ref{Fig_phantom_result}. The probe \revision{reached} the desired orientation at $8~s$ and then, the optimal orientation is maintained until the end of the sweep at $16~s$. This shows that the proposed method can perform automatic US screening based on real-time US imaging. 
\revision{In addition, it is noteworthy that false-positive cases of real-time segmentation may occur during scanning (e.g., Fig.~\ref{Fig_phantom_result} (b)). To correctly select a good candidate in such a case, the distances between the selected vessel in the previous frame and the detected results in the new frame were calculated. Since the vessel is a continuous object, the detected result with the smallest distance is finally selected to reduce the chance of tracking the wrong object.}
To analyze whether the proposed method is able to run in real-time, the image processing and optimization overhead were measured. \revision{The entire process took $32~ms$ on average while the U-Net only took $5~ms$ for image processing (TABLE~\ref{Table_segmentation_result}}).
Therefore, the system can be run in real-time. 



\section{Discussion}
This work enables RUSS to automatically perform US sweeps for tubular structures without any requirement for pre-operative images. The proposed approach is able to optimize the probe pose (position and orientation) and calculate vessel radius in real-time during US sweeps. Since autonomous RUSS overcomes the limitation of human hand-eye coordination, a stable and accurate prediction of the radius can be achieved. Significantly, due to the use of multiple frames, the proposed optimization approach is able to accurately calculate the vessel radius when the US probe is titled from normal direction.
\secondrevision{Compared with existing methods, the approach also demonstrated the advantage of optimizing multiple objects meanwhile. The errors of the probe orientation, target centering and estimated radius are mainly caused by the error of the image segmentation results. In addition, the difference between the real orientation error and estimated orientation error is because of the compliant behaviour of the controller. To guarantee patients' safety and good contact condition between probe and object surface, the impedance control was employed, however, this controller allows positional error to achieve compliant performance.  
}

In addition, \thirdrevision{the main limitation of this work is that the segmentation network was only tested on the brachial artery in the forearm. A more generic model for segmenting different arteries has to be developed to further test the proposed approach in different clinical trials.} Considering variations between patients, to make a network accurately segment target tissues for most people, Prevost~\emph{et al.} suggested that several tens of US sweeps with different acquisition parameters on several tens of volunteers are required to be used for training~\cite{prevost20183d}. Alternatively, to overcome the difficulties in collecting a larger number of in-vivo data sets, large-scale simulated US data and small in-vivo data sets can be used together to train an generic segmentation network, as in~\cite{patel2020improved}. Beside, the technique for correcting the pressure-induced deformation in US images~\cite{jiang2021motion} can be further integrated to generate accurate 3D anatomy.




\section{Conclusion}
In this work, we introduced an end-to-end workflow for autonomous robotic screening of tubular structures based only on real-time US images. The proposed method can automatically adjust the probe position to horizontally center a tracked object and tune the probe orientation to the normal direction of the target vessel during US sweeps. Additionally, the radius of the target object is calculated in real-time, even when the probe is titled from the normal direction. 
The results demonstrated that the proposed method is able to automatically perform US sweeps and accurately estimate the geometry of the target vessels both in simulation ($e_{ra}$: $1.16\pm0.1~mm$, $e_{or}$: $2.7\pm3.3^{\circ}$) and for a real phantom ($e_{ra}$: $0.77\pm0.4~mm$, $e_{or}^{rea}$: $3.7\pm1.6^{\circ}$, $e_{ce}$: $0.24\pm0.19~mm$). 
The development of autonomous RUSS is a promising way to overcome inter-operator variability providing repeatable US images. In addition, with such a RUSS, sonographers can focus on diagnosis to fully utilize their experience and knowledge. This approach could be integrated with autonomous diagnosis techniques to further pave the way for a fully automatic US-guided intervention system~\cite{guo2019novel}. 




\section*{ACKNOWLEDGMENT}
\revision{
The authors would like to acknowledge the Editor-In-Chief, Associate Editor, and anonymous reviewers for their contributions to the improvement of this article. Besides, Z. Jiang wants to thank, in particular, the invaluable supports from his wife B. Zhang, the hero who just gave birth to their lovely daughter E. M. Jiang on 14.06.2021.}


\bibliographystyle{IEEEtran}
\bibliography{IEEEabrv,references} 

\begin{IEEEbiography}[{\includegraphics[width=1in,height=1.25in,clip,keepaspectratio]{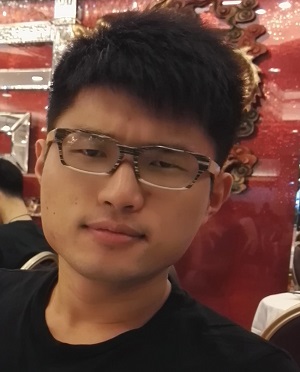}}]
{Zhongliang Jiang} (Graduate Student Member, IEEE) received the M.Eng. degree in Mechanical Engineering from the Harbin Institute of Technology, Shenzhen, China, in 2017. From January 2017 to July 2018, he worked as a research assistant in the Shenzhen Institutes of Advanced Technology (SIAT) of the Chinese Academy of Science (CAS), Shenzhen, China. He is currently working toward the Ph.D. degree in Computer Science with the Technical University of Munich, Munich, Germany.

His research interests include medical robotics, robotic learning, human-robot interaction, and robotic ultrasound.
\\
\end{IEEEbiography}

\vspace{-1cm}
\begin{IEEEbiography}[{\includegraphics[width=1in,height=1.25in,clip,keepaspectratio]{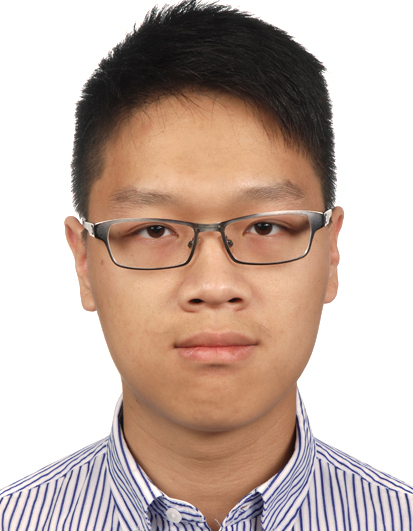}}]
{Zhenyu Li} received the B.Sc. degree and the M.Sc. degree in Electrical Engineering and Information Technology from the Technical University of Ilmenau, Ilmenau, Germany, in 2018, and the Technical University of Munich, Munich, Germany, in 2021, respectively.

His research interests include robot manipulation, robotic learning, and computer vision.
\\ 
\end{IEEEbiography}

\newpage
\vspace{-1cm}
\begin{IEEEbiography}[{\includegraphics[width=1in,height=1.25in,clip,keepaspectratio]{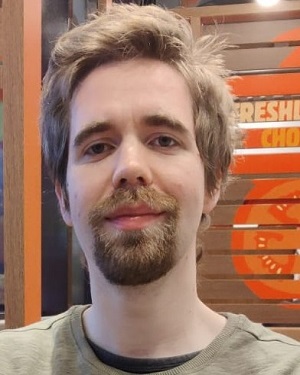}}]
{Matthias Grimm} received the B.Sc. and M.Sc. in Computer Science in 2015 and 2018, respectively, from the Technical University of Munich, Munich, Germany, where he is working toward the Ph.D. degree in Computer Science. 

His research interests include image registration and 3D reconstruction.
\\ 
\\ 
\end{IEEEbiography}

\vspace{-1cm}
\begin{IEEEbiography}[{\includegraphics[width=1in,height=1.25in,clip,keepaspectratio]{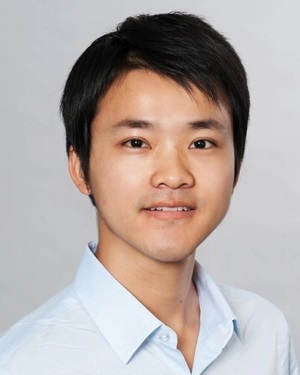}}]
{Mingchuan Zhou} received Ph.D. degree in computer science from the Technical University of Munich, Munich, Germany, in 2020. He was the visiting scholar at Laboratory for Computational Sensing and Robotics, Johns Hopkins University, USA, 2019. He was a joint postdoc at Institute of Biological and Medical Imaging (IBMI) of the Helmholtz Center Munich and Chair for Computer Aided Medical Procedures Augmented Reality (CAMP) at the Technical University of Munich from 2019 to 2021. He is currently an Assistant Professor leading multi scale robotic manipulation lab for agriculture in Zhejiang University. His research interests include the autonomous system, agricultural robotics, medical robotics, and image processing. 
 
Dr. Zhou was the recipient of Finalist of Best Paper Awards IEEE ROBIO in 2017 and best poster award in IEEE ICRA 2021 workshop of Task-Informed Grasping: Agri-Food manipulation (TIG-III).


\\ 
\end{IEEEbiography}

\vspace{-1cm}
\begin{IEEEbiography}[{\includegraphics[width=1in,height=1.25in,clip,keepaspectratio]{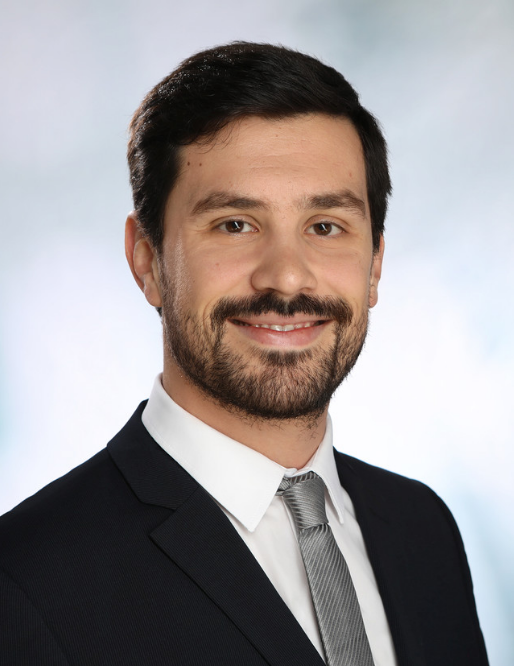}}]
{Marco Esposito} holds a PhD on Collaborative Robotic Medical Imaging and a Master's degree in Computer Science from the Technical University of Munich, Munich, Germany. Previously, he obtained B.Sc. degree in Computer Engineering from the University of Pisa. During his academic career he conjugated research and contribution to the open source robotic community. His work was presented at top-tier conferences such as MICCAI and ROSCON, and received the IJCARS Best Paper Award. In May 2019, Marco joined ImFusion, where he supports the translation of state-of-the-art methods in freehand ultrasound imaging and robotics into innovative medical devices.
\\ 
\end{IEEEbiography}

\vspace{-1cm}
\begin{IEEEbiography}[{\includegraphics[width=1in,height=1.25in,clip,keepaspectratio]{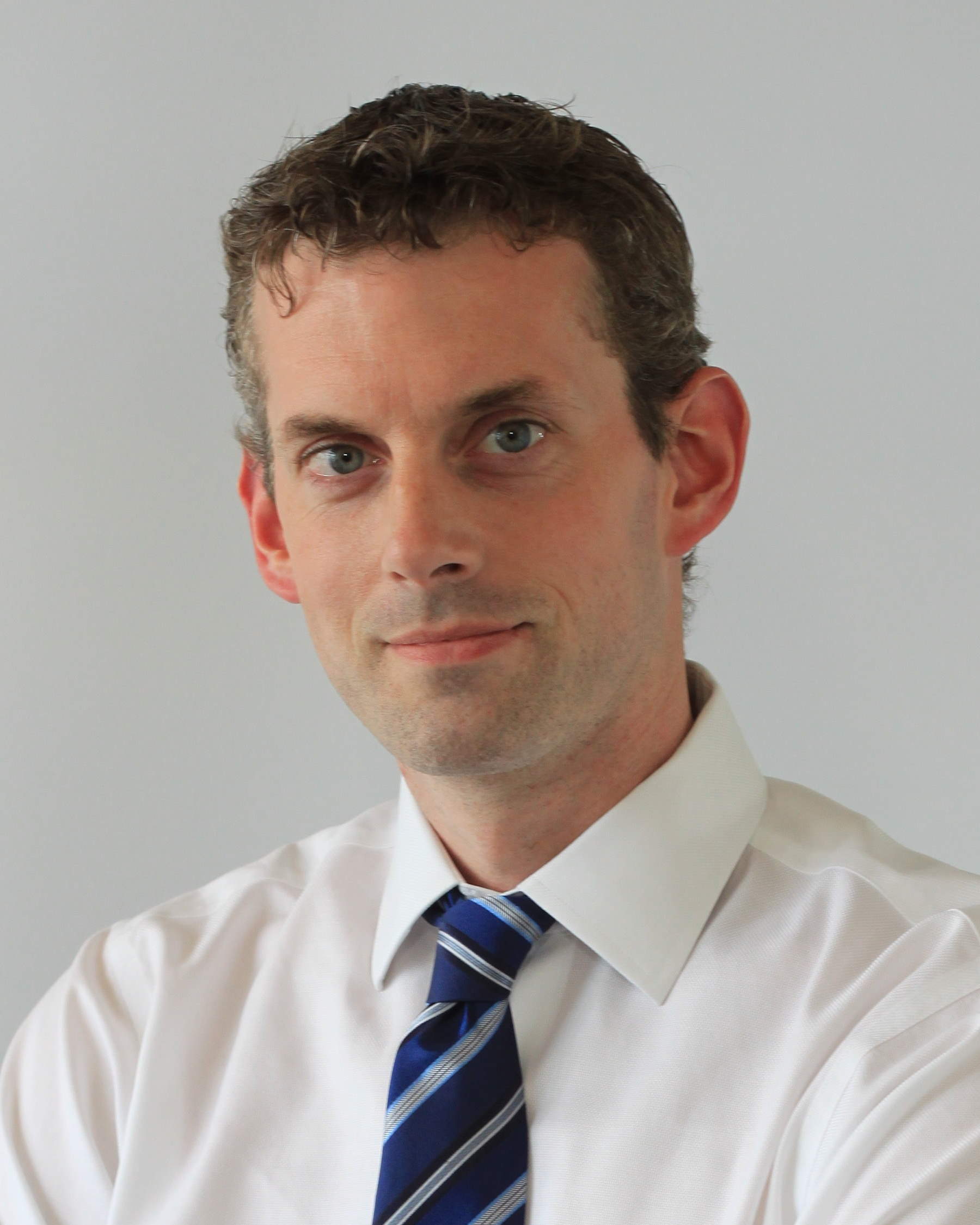}}]
{Wolfgang Wein} received the Ph.D. degree in Computer Science from the Technical University of Munich, Germany, in 2007. He worked a few years at Siemens Corporate Research in Princeton, NJ USA.

He is currently the founder and CEO of the medical R\&D company ImFusion. He has conducted numerous projects from early feasibility to product implementation, providing innovative solutions for various clinical applications. From receiving the MICCAI Young Scientist Award in 2007, he has remained actively involved in the academic community with regular publications, and also has a teaching assignment for the faculty of computer science at TU Munich.
\\ 
\end{IEEEbiography}

\vspace{-1cm}
\begin{IEEEbiography}[{\includegraphics[width=1in,height=1.25in,clip,keepaspectratio]{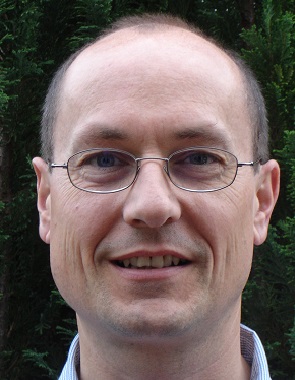}}]
{Walter Stechele} received the Dipl.-Ing. and Dr.-Ing. degrees in electrical engineering from the Technical University of Munich, Germany, in 1983 and 1988, respectively. In 1990 he joined Kontron Elektronik GmbH, a German electronic company, where he was responsible for the ASIC and PCB design department. Since 1993 he has been Academic Director at the Technical University of Munich. 

His interests include visual computing and robotic vision, with focus on Multi Processor System-on-Chip (MPSoC). 
\\ 
\end{IEEEbiography}

\vspace{-1cm}
\begin{IEEEbiography}[{\includegraphics[width=1in,height=1.25in,clip,keepaspectratio]{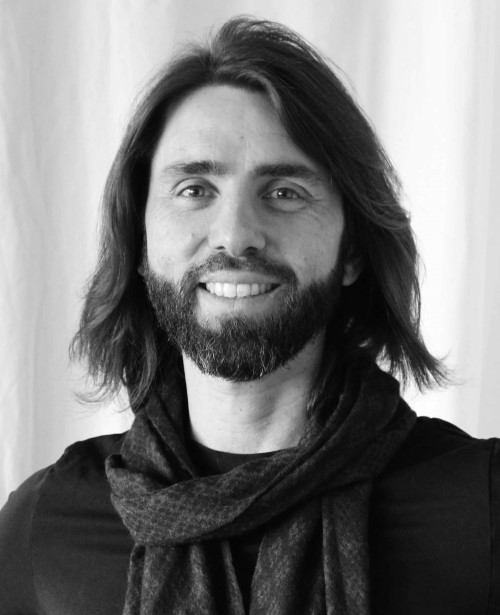}}]
{Thomas Wendler} received the M.Sc. degree in Biomedical Engineering and the Ph.D. degree in Computer Science from the Technical University of Munich, Germany. He is currently the Vice-Director of the chair for Computer-Aided Medical Procedures and Augmented Reality (CAMP) at the Technical University of Munich (TUM), CEO of ScintHealth GmbH, and CTO of SurgicEye GmbH.

His main research focus is the translation of novel computer-aided intervention tools and data-driven medical image analysis methods for clinical decision support into clinical applications. 

\\ 
\end{IEEEbiography}

\vspace{-1cm}
\begin{IEEEbiography}[{\includegraphics[width=1in,height=1.25in,clip,keepaspectratio]{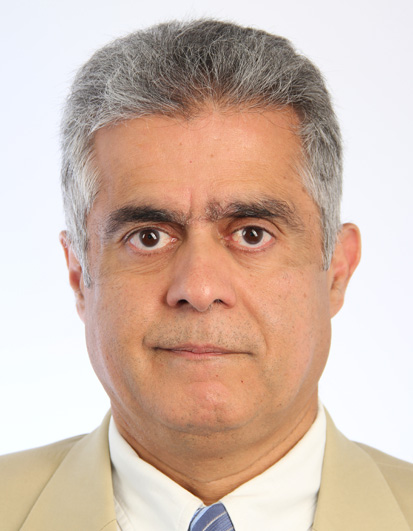}}]
{Nassir Navab} received the Ph.D. degree in computer and automation with INRIA, and the University of Paris XI, Paris, France, in 1993.

He is currently a Full Professor and the Director of the Laboratory for Computer-Aided Medical Procedures with the Technical University of Munich, Munich, Germany, and with the Johns Hopkins University, Baltimore, MD, USA. He has also secondary faculty appointments with the both affiliated Medical Schools. He enjoyed two years of a Postdoctoral Fellowship with the MIT Media Laboratory, Cambridge, MA, USA, before joining Siemens Corporate Research (SCR), Princeton, NJ, USA, in 1994. 

Dr. Navab was a Distinguished Member and was the recipient of the Siemens Inventor of the Year Award in 2001, at SCR, the SMIT Society Technology award in 2010 for the introduction of Camera Augmented Mobile C-arm and Freehand SPECT technologies, and the ``$10$ years lasting impact award" of IEEE ISMAR in 2015. In 2012, he was elected as a Fellow of the MICCAI Society. He is the author of hundreds of peer-reviewed scientific papers, with more than 42,500 citations and an h-index of 92 as of June 23, 2021. He is the author of more than thirty awarded papers including 11 at MICCAI, 5 at IPCAI, and three at IEEE ISMAR. He is the inventor of 50 granted US patents and more than 50 International ones.
\\ \\ 
\end{IEEEbiography}




\end{document}